\documentclass[10pt,twocolumn,letterpaper]{article}

\usepackage{cvpr}              

\usepackage{graphicx}
\usepackage{amsmath}
\usepackage{amssymb}
\usepackage{booktabs}
\usepackage[accsupp]{axessibility}

\usepackage{algorithm, algorithmic}
\renewcommand{\algorithmicrequire}{\textbf{Input:}}  
\renewcommand{\algorithmicensure}{\textbf{Output:}} 

\usepackage[pagebackref,breaklinks,colorlinks]{hyperref}

\usepackage[capitalize]{cleveref}
\crefname{section}{Sec.}{Secs.}
\Crefname{section}{Section}{Sections}
\Crefname{table}{Table}{Tables}
\crefname{table}{Tab.}{Tabs.}

\def\cvprPaperID{1378} 
\def\confName{CVPR}
\def\confYear{2023}

\begin{document}

\title{Not All Image Regions Matter: \\
Masked Vector Quantization for Autoregressive Image Generation}

\author{Mengqi Huang\textsuperscript{\rm 1}, Zhendong Mao\textsuperscript{\rm 1, 3}\thanks{Zhendong Mao is the corresponding author.}, Quan Wang\textsuperscript{\rm 2}, Yongdong Zhang\textsuperscript{\rm 1, 3} \\
\textsuperscript{\rm 1}University of Science and Technology of China,
Hefei, China;
\textsuperscript{\rm 2}MOE Key Laboratory of Trustworthy, \\
Distributed Computing and Service, Beijing University of Posts and Telecommunications, Beijing, China; \\
\textsuperscript{\rm 3}Institute of Artificial intelligence, Hefei Comprehensive National Science Center, Hefei, China \\
{\tt\small huangmq@mail.ustc.edu.cn, \{zdmao, zhyd73\}@ustc.edu.cn, wangquan@bupt.edu.cn}
}
\maketitle

\begin{abstract}
   Existing autoregressive models follow the two-stage generation paradigm that first learns a codebook in the latent space for image reconstruction and then completes the image generation autoregressively based on the learned codebook. However, existing codebook learning simply models all local region information of images without distinguishing their different perceptual importance, which brings redundancy in the learned codebook that not only limits the next stage’s autoregressive model’s ability to model important structure but also results in high training cost and slow generation speed. In this study, we borrow the idea of importance perception from classical image coding theory and propose a novel two-stage framework, which consists of Masked Quantization VAE (MQ-VAE) and Stackformer, to relieve the model from modeling redundancy. Specifically, MQ-VAE incorporates an adaptive mask module for masking redundant region features before quantization and an adaptive de-mask module for recovering the original grid image feature map to faithfully reconstruct the original images after quantization. Then, Stackformer learns to predict the combination of the next code and its position in the feature map. Comprehensive experiments on various image generation validate our effectiveness and efficiency. Code will be released at \url{https://github.com/CrossmodalGroup/MaskedVectorQuantization}.
\end{abstract}

\section{Introduction}
\label{sec:intro}

Deep generative models of images have received significant improvements over the past few years and broadly fall into two categories: likelihood-based models, which include VAEs\cite{kingma2013auto}, flow-based\cite{rezende2015variational}, diffusion models\cite{ho2020denoising} and autoregressive models\cite{van2016pixel}, and generative adversarial networks (GANs)\cite{goodfellow2014generative}, which use discriminator networks to distinguish samples from generator networks and real examples. Compared with GANs, likelihood-based models' training objective, \emph{i.e.}, the negative log-likelihood (NLL) or its upper bound, incentives learning the full data distribution and allows for detecting overfitting. 

\begin{figure}
  \centering
  \includegraphics[width=1.0\linewidth]{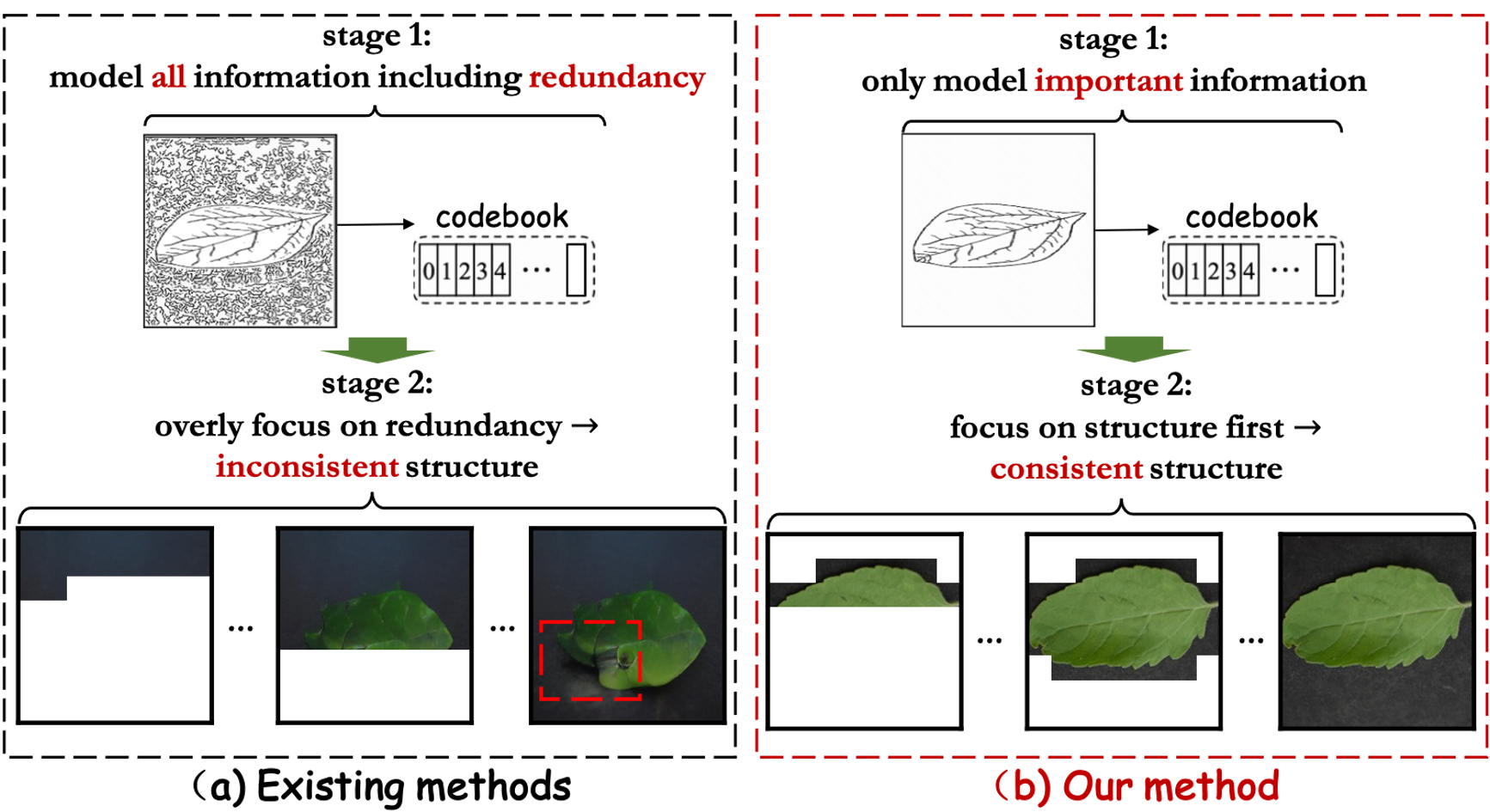}
  \caption{Illustration of our motivation. (a) Existing works model all local regions without distinguishing their perceptual importance in stage 1, which not only brings redundancy (\emph{e.g.}, the textural regions like the background) in the learned codebook but also make the autoregressive models overly focus on modeling this redundancy and hinder other important structural regions modeling. (b) The codebook learning in our method only includes the important regions, \emph{e.g.}, the structural regions like corners and edges, since other unimportant ones can be restored even if missing, and thus autoregressive model could focus on modeling these important regions in stage 2 and results in better generation quality.}
  \label{motivation}
\end{figure}

Among the likelihood-based models, autoregressive models have recently attracted increasing attention for their impressive modeling ability and scalability. Recent autoregressive image generation\cite{van2017neural,ding2021cogview, esser2021imagebart, esser2021taming, lee2022autoregressive, ramesh2021zero, razavi2019generating,lee2022autoregressive,shin2021translation} follows the two-stage generation paradigm, \emph{i.e.}, the first stage learns a codebook in the latent space for image reconstruction and the second stage completes the image generation in the raster-scan\cite{esser2021taming} order by autoregressive models based on the learned codebook. Since codebook learning in the first stage defines the discrete image representation for the next autoregressive modeling, a high-quality codebook is the key to generate high-quality images. Several recent works focus on improving the codebook learning in the first stage, \emph{e.g.}, VQGAN\cite{esser2021taming} introduces adversarial loss and perceptual loss. ViT-VQGAN\cite{yu2021vector} introduces a more expressive transformer backbone. RQ-VAE\cite{lee2022autoregressive} introduces the residual quantization to reduce the resolution of the latent space. In general, the essence of existing codebook learning is the modeling of all local region information (\emph{i.e.}, an $8\times8$ or $16\times16$ patch) of images in the dataset, without distinguishing their different perceptual importance.

In this study, we point out that existing codebook learning exists gaps with classical image coding theory\cite{kunt1985second,jayant1993signal,kocher1982contour}, the basic idea of which is to remove redundant information by perceiving the importance of different regions in images. The image coding theory reveals that an ideal image coding method should only encode images’ perceptually important regions (\emph{i.e.}, which cannot be restored if missing) while discarding the unimportant ones (\emph{i.e.}, which can be restored by other image regions even if missing). The neglect of considering such perceptual importance in existing works poses problems in two aspects, as illustrated in Figure \ref{motivation}(a): (1) the existence of this large amount of repetitive and redundant information brings redundancy to the learned codebook, which further makes the autoregressive model in the next stage overly focus on modeling this redundancy while overlooking other important regions and finally degrades generation quality. (2) the redundancy makes the autoregressive model need to predict more (redundant) quantized codes to generate images, which significantly increases the training cost and decreases the generating speed. Although the effectiveness and efficiency of image coding theory have been widely validated, how to introduce this idea into codebook learning remains unexplored.

The key of applying image coding theory to codebook learning is to distinguish important image parts from unimportant ones correctly. Considering that the essential difference between these two sets lies in whether they can be restored if missing, we found that this distinction can be realized through the mask mechanism, \emph{i.e.}, the masked part is important if it cannot be faithfully restored, and otherwise unimportant. Based on the above observation, we thereby propose a novel two-stage generation paradigm upon the mask mechanism to relieve the model from modeling redundant information. Specifically, we first propose a \emph{Masked Quantization VAE (MQ-VAE)} with two novel modules, \emph{i.e.}, an \emph{adaptive mask module} for adaptively masking redundant region features before quantization, and an \emph{adaptive de-mask module} for adaptively recovering the original grid image feature map to faithfully reconstruct original images after quantization. As for the \emph{adaptive mask module}, it incorporates a lightweight content-aware scoring network that learns to measure the importance of each image region feature. The features are then ranked by the importance scores and only a subset of high-scored features will be quantized further. As for the \emph{adaptive de-mask module}, we design a direction-constrained self-attention to encourage the information flow from the unmasked regions to the masked regions while blocking the reverse, which aims to infer the original masked region information based on unmasked ones. Thanks to the adaptive mask and de-mask mechanism, our MQ-VAE removes the negative effects of redundant image regions and also shortens the sequence length to achieve both effectiveness and efficiency.

Moreover, since different images have different important regions, the position of quantized codes in the feature map also dynamically changed. Therefore, we further propose \emph{Stackformer} for learning to predict the combination of both codes and their corresponding positions. Concretely, the proposed \emph{Stackformer} stacks a Code-Transformer and a Position-Transformer, where the Code-Transformer learns to predict the next code based on all previous codes and their positions, and the Position-Transformer learns to predict the next code’s position based on all previous codes’ positions and \emph{current} code. 

With our method, as shown in Figure \ref{motivation}(b), the codebook learning only includes the important regions, \emph{e.g.}, the structural regions, since unimportant ones like the background can be restored even if missing. And therefore the autoregressive model in the second stage could focus on modeling these important regions and brings better generation quality.

In a nutshell, we summarize our main contributions as:

\textcolor{blue}{\textbf{Conceptually}}, we point out that existing codebook learning ignores distinguishing the perceptual importance of different image regions, which brings redundancy that degrades generation quality and decreases generation speed.

\textcolor{blue}{\textbf{Technically}}, (i) we propose MQ-VAE with a novel \emph{adaptive mask module} to mask redundant region features before quantization and a novel \emph{adaptive de-mask module} to recover the original feature map after quantization; (ii) we propose a novel \emph{Stackformer} to predict the combination of both codes and their corresponding positions.

\textcolor{blue}{\textbf{Experimentally}}, comprehensive experiments on various generations validate our effectiveness and efficiency, \ie, we achieve 8.1\%, 2.3\%, and 18.6\% FID improvement
on un-, class-, and text-conditional state-of-the-art at million-level parameters, and faster generation speed
compared to existing autoregressive models.





\begin{figure*}
  \centering
  \includegraphics[width=1.0\linewidth]{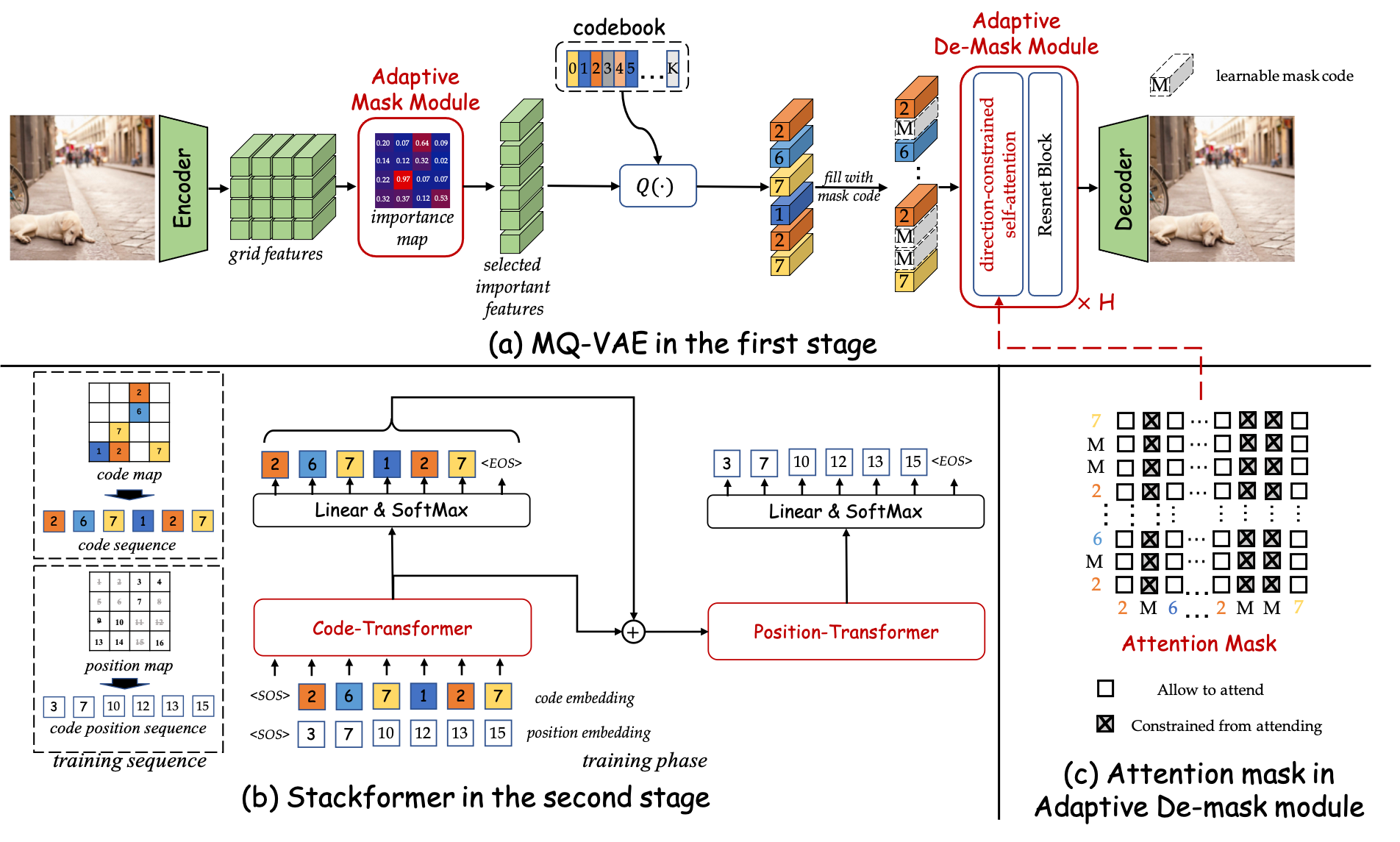}
  \caption{The Illustration of our proposed two-stage generation framework. (a) In the first stage, MQ-VAE adaptively masks the redundant region features to prevent redundant codes while keeping important ones, which ensures that the original images can still be faithfully recovered. 
  Here, $\frac{10}{16}$ regions are masked and $\frac{6}{16}$ regions are kept. 
  (b) In the second stage, Stackformer stacks a Code-Transformer and a Position-Transformer to autoregressively predict the next code and its position in the original 2D feature map, respectively. 
  (c) The attention mask of the proposed direction-constrained self-attention in the adaptive de-mask module for inferring masked regions features.}
  \label{framework}
\end{figure*}

\section{Related Work}
\subsection{Autoregressive Modeling for Image Generation}

Autoregressive models for image generation have recently attracted increasing research attention and have shown impressive results\cite{ramesh2021zero, ding2021cogview, esser2021imagebart, esser2021taming, van2017neural, yu2021vector, razavi2019generating, nash2021generating, shin2021translation} among various generation tasks. Early autoregressive models\cite{van2016pixel, parmar2018image, chen2020generative} directly optimizing the likelihood of raw image pixels, \emph{e.g.}, Image-GPT\cite{chen2020generative} trains a transformer\cite{vaswani2017attention} to autoregressively predict pixels' cluster centroids, which could only generate images with a maximum resolution of $64\times64$. \cite{van2017neural} presents the Vector Quantized Variational Autoencoder (VQVAE), which learns images' low-dimension discrete representation and models their distribution autoregressively. VQ-VAE2\cite{razavi2019generating} extends this approach using a hierarchy of discrete representations. VQGAN\cite{esser2021taming} further improves the perceptual quality of reconstructed images using adversarial\cite{goodfellow2014generative, isola2017image} and perceptual loss\cite{ledig2017photo}. ViT-VQGAN\cite{yu2021vector} introduces a more expressive transformer backbone. RQ-VAE\cite{lee2022autoregressive} uses Residual Quantization\cite{juang1982multiple, martinez2014stacked} to iteratively quantizes a vector and its residuals and represent the vector as a stack of tokens. Although vector quantization has become the fundamental technique for modern visual autoregressive models, the critical removing redundancy in codebook learning has not been explored yet, which becomes a critical bottleneck.

\subsection{Masked Modeling} 
Masked modeling is popular among both natural language processing and computer vision. BERT\cite{devlin2018bert} randomly masks a portion of the input sequence and trains models to predict the missing content. In the computer vision domain, the ViT\cite{dosovitskiy2020image} studies masked patch prediction for self-supervised learning. BEiT\cite{bao2021beit} proposes to predict discrete tokens. Most recently, MaskGIT\cite{chang2022maskgit} also used the masking strategy for VQ-based image generation. However, our proposed method differs from MaskGIT in two aspects: (1) Our primary motivation for the masking strategy applied in the proposed MQ-VAE in the first stage aims to learn a more compact and effective vector quantization (VQ) itself by masking perceptual unimportant regions, while MaskGIT uses masking strategy in the second stage to better use a learned VQ. (2) The mask in our proposed MQ-VAE is learned and adaptively changed according to different image content, while the mask in MaskGIT is randomly sampled for the mask-and-predict training. In conclusion, to the best of our knowledge, this is the first time that masked modeling has been applied for vector quantization.

\section{Methodology}

We propose a novel two-stage framework with MQ-VAE and Stackformer for autoregressive image generation, as illustrated in Figure \ref{framework}. MQ-VAE only masks redundant region features to prevent redundant codes and Stackformer stacks two transformers to autoregressively predict the next code and its position. In the following, we will first briefly revisit the formulation of vector quantization and then describe our proposed method in detail.

\subsection{Preliminary}

We follow the definition and notations of previous works\cite{esser2021taming ,lee2022autoregressive}. Specifically, the \emph{codebook} $\mathcal{C} := \{(k, \boldsymbol{e}(k))\}_{k \in [K]}$ is defined as the set of finite pairs of code $k$ and its code embedding $\boldsymbol{e}(k) \in \mathbb{R}^{n_z}$. Here $K$ is the codebook size and $n_z$ is the code dimension. An image $\mathbf{X} \in \mathbb{R}^{H_0 \times W_0 \times 3}$ is first encoded into grid features $\mathbf{Z} = E(\mathbf{X}) \in \mathbb{R}^{H \times W \times n_z}$ by the encoder $E$, where $(H, W) = (H_0/f, W_0/f)$ and $f$ is the corresponding downsampling factor. For each vector $\boldsymbol{z} \in \mathbb{R}^{n_z}$ in $\mathbf{Z}$, it is replaced with the code embedding that has the closest euclidean distance with it in the codebook $\mathcal{C}$ through the vector quantization operation $\mathcal{Q(\cdot)}$:


\begin{equation}
    \mathcal{Q}(\boldsymbol{z}; \mathcal{C}) = \arg\min\limits_{k \in [K]} || \boldsymbol{z} - \boldsymbol{e}_k ||^{2}_{2}.
\end{equation}
Here, $\mathcal{Q}(\boldsymbol{z}; \mathcal{C})$ is the quantized code. $\boldsymbol{z^q} = \boldsymbol{e}(\mathcal{Q}(\boldsymbol{z}; \mathcal{C}))$ is the quantized vector. By applying $\mathcal{Q(\cdot)}$ to each feature vector, we could get the quantized code map $\mathbf{M} \in [K]^{H \times W}$ and the quantized features $\mathbf{Z}^{\boldsymbol{q}} \in \mathbb{R}^{H \times W \times n_z}$. The original image is reconstructed by the decoder $D$ as $\Tilde{\mathbf{X}} = D(\mathbf{Z}^{\boldsymbol{q}})$.


\subsection{Stage 1: MQ-VAE}

Existing methods quantize each feature vector of $\mathbf{Z}$ without distinguishing their different perceptual importance and thus bring redundancy in the learned codebook, which not only degrades the generation quality but also decreases the generation speed. To relieve the model from this redundancy, we propose MQ-VAE with two novel modules, \emph{i.e.}, the \emph{adaptive mask module} for adaptively masking redundant region features before vector quantization and \emph{adaptive de-mask module} for adaptively recovering the original grid image feature map after vector quantization. 

\textbf{Adaptive Mask Module.} The encoded grid feature map $\mathbf{Z} \in \mathbb{R}^{H \times W \times n_e}$ is first flattened into $\mathbf{Z} \in \mathbb{R}^{L \times n_e}$, where $L = H \times W$. The proposed adaptive mask module then uses a lightweight scoring network $f_{s}$ to measure the importance of each region feature $\boldsymbol{z}_{l}$ in $\mathbf{Z}$, which is implemented as a two-layer MLP:

\begin{equation}
    s_{l} = f_{s}(\boldsymbol{z}_{l}), l=1, ..., L.
\end{equation}
The larger score $s_{l}$ is, the more important the region feature $\boldsymbol{z}_{l}$ is. Then the region features are sorted in descending order according to the predicted scores. The sorted region features and their scores are denoted as $\{\boldsymbol{z}^{'}_{l} \}$ and $\{s^{'}_{l} \}$ respectively, where $l=1, ..., L$. To enable the learning of $f_{s}$, the predicted scores are further multiplied with the normalized region features as modulating factors. We select the top $N$ scoring vectors as the important region features, 

\begin{gather}
    \hat{\mathbf{Z}} = \{\boldsymbol{z}^{''}_l | \boldsymbol{z}^{''}_l = \text{LayerNorm}(\boldsymbol{z}^{'}_l)*s^{'}_l\},  l=1, ..., N.\\ \hat{\mathbf{P}} = \{p_{z_l^{''}} | p_{z_l^{''}} \in \{ 0, ..., L\}\},  l=1, ..., N.
\end{gather}
Here, $\hat{\mathbf{Z}}$ denotes the selected important region features set, and $\hat{\mathbf{P}}$ denotes the corresponding position set that represents the position of each selected region feature in the original 2D feature map. The selected number $N = \alpha \times L$, where $\alpha$ is a constant fractional value. The mask ratio is defined as $1 - \alpha$. This design also enables a flexible trade-off between the image generation speed and image generation quality, which we will discuss in experiments. After obtaining $\hat{\mathbf{Z}}$, we further apply the quantization function $\mathcal{Q}$ to each of them and obtain the quantized important region features set $\hat{\mathbf{Z}}^{\boldsymbol{q}}$ as well as its code matrix $\hat{\mathbf{M}}$.

\textbf{Adaptive De-mask Module.} After quantization, we fill the quantized features $\hat{\mathbf{Z}}^{\boldsymbol{q}}$ back into the original 2D feature map according to $\hat{\mathbf{P}}$,  while other masked positions are filled with a uniformly initialized learnable mask code embedding, as shown in Figure \ref{framework}(a). Directly inputting filled grid features to the decoder $D$ could bring sub-optimal reconstruction results since the mask code embedding here only serves as the placeholders that contain little information. Therefore, we further propose the adaptive de-mask module, which applies a novel \emph{direction-constrained self-attention} to encourage the information flow from unmasked regions to the masked ones while blocking the reverse. Such a design allows the model to utilize the unmasked region features to infer the masked ones while also preventing the masked regions to have negative impacts on the unmasked ones since they are less informative.

Our adaptive de-mask module is implemented as $H$ identical sub-modules, where each consists of a direction-constrained self-attention block and a Resnet block. The direction-constrained self-attention is mathematically formed as (Resnet block is omitted for simplicity):

\begin{gather}
    \boldsymbol{q}, \boldsymbol{k}, \boldsymbol{v} = \boldsymbol{W}_{q} \hat{\mathbf{Z}}^{\boldsymbol{q}, h}, \boldsymbol{W}_{k} \hat{\mathbf{Z}}^{\boldsymbol{q}, h}, \boldsymbol{W}_{v} \hat{\mathbf{Z}}^{\boldsymbol{q}, h}\\
    \boldsymbol{A} = (\text{SoftMax}(\frac{\boldsymbol{q}\boldsymbol{k}^T}{\sqrt{n_e}})) \odot \boldsymbol{B_h} \\
    \hat{\mathbf{Z}}^{\boldsymbol{q}, h+1} = \boldsymbol{A}\boldsymbol{v}.
\end{gather}

Here $h \in \{ 1, .., H\}$ and $\hat{\mathbf{Z}}^{\boldsymbol{q}, 1}$ is the filled quantized grid features. $\boldsymbol{W}_{q}, \boldsymbol{W}_{k}, \boldsymbol{W}_{v} \in \mathbb{R}^{n_e \times n_e}$ are the learnable parameters. $\boldsymbol{B}_h$ is the attention mask at $h$ sub-module. Specifically, since the initial mask code contains little information, we define $\boldsymbol{B}_1 = [b_l \in \{ 0, 1\}, | l =1, ..., L] \in \mathbb{R}^{1 \times L}$ to forbid it from attending to other unmasked codes to avoid negative impact, where $0$ for the mask position and $1$ for the unmasked. Considering that the mask code is updated with more and more information after each sub-module, we propose to synchronously amplify its interaction with other codes step by step through a mask updating mechanism:

\begin{equation}
    \boldsymbol{B}_{h+1} = \sqrt{\boldsymbol{B}_h},
\end{equation}
where the initial $0$ in $\boldsymbol{B}_1$ is replaced with a small number 0.02, since $\sqrt 0$ will always result in 0. Finally, the original image is recovered by the decoder as $\Tilde{\mathbf{X}} = D(\hat{\mathbf{Z}}^{\boldsymbol{q}, H})$.

\subsection{Stage 2: Stackformer}
The perceptual important regions of different images vary. Therefore the positions of quantized codes in the feature map also dynamically change along with the image content.  As a result, our proposed MQ-VAE formulates an image as both the code sequence $\hat{\mathbf{M}}$ and the code position sequence $\hat{\mathbf{P}}$. To simultaneously learn the combination of the codes and their positions, we propose a novel Stackformer, which stacks a Code-Transformer and a Position-Transformer. The Code-Transformer learns to predict the next code based on all previous steps' codes and their positions, while the Position-Transformer learns to predict the next code's position based on all previous steps' positions and current code. Directly treating the importance descending order sequence $\hat{\mathbf{M}}$ and $\hat{\mathbf{P}}$ as the inputs are natural, but the dramatic position changes of adjacent code could make the network hard to converge. For example, the position of the first code may be in the upper left corner of the image, while the position of the second code may be in the lower right corner of the image. Therefore, we further propose to use the raster-scan order\cite{esser2021taming} to rearrange both sequences to deal with the converge problem.

Mathematically, taking the raster-scan code and code position sequence $(\overline{\mathbf{M}}, \overline{\mathbf{P}}) = \text{rearrange}(\hat{\mathbf{M}}, \hat{\mathbf{P}})$, Stackformer learns $p(\overline{\mathbf{M}}, \overline{\mathbf{P}})$, which is autoregressively factorized as:

\begin{equation}
    p(\overline{\mathbf{M}}, \overline{\mathbf{P}}) = \prod \limits_{l=1}^N 
    p(\overline{\mathbf{M}}_l | \overline{\mathbf{M}}_{<l}, \overline{\mathbf{P}}_{<l})
    p(\overline{\mathbf{P}}_{l} | \overline{\mathbf{M}}_{\leq l}, \overline{\mathbf{P}}_{<l})
\end{equation}

\textbf{Code-Transformer} takes the sum of code embeddings $\boldsymbol{e}_c(\cdot)$ and code position embedding $\boldsymbol{e}_{p}(\cdot)$ as inputs:

\begin{equation}
    \mathbf{U}_c = \boldsymbol{e}_c(\overline{\mathbf{M}}_{[1:N_c+N]}) + \boldsymbol{e}_{p1}(\overline{\mathbf{P}}_{[1:N_c+N]}),
\end{equation}
where $N_c$ is the condition length. For the unconditional generation, we add a $<$sos$>$ code at the start of the code and code position sequence. For conditioning, we append class or text codes to the start of the code sequence and the same length of $<$sos$>$ code to the code position sequence. We further add an extra learned absolute position embedding to $\mathbf{U}_c$ to form the final input, which makes the network aware of the absolute position of the sequence. After processing by Code-Transformer, the output hidden vector $\mathbf{H}_c$ encodes both code and their position information and is used for the next code prediction. The negative log-likelihood (NLL) loss for code autoregressive training is:

\begin{equation}
    \mathcal{L}_{\text{code}} = \mathbb{E}[-\log p(\overline{\mathbf{M}}_l |\overline{\mathbf{M}}_{<l}, \overline{\mathbf{P}}_{<l}]).
\end{equation}

\textbf{Position-Transformer} takes the sum of Code-Transformer's output hidden vector $\mathbf{H}_c$ and an extra code embedding as input:

\begin{equation}
    \mathbf{U}_p = \mathbf{H}_c[N_c: N_c+N-1] + \boldsymbol{e}_c(\overline{\mathbf{M}}_{[N_c+1:N_c+N]}).
\end{equation}
Here $\mathbf{U}_p$ is the input for Position-Transformer and the information of current code is included in $\boldsymbol{e}_c(\overline{\mathbf{M}}_{[N_c+1:N_c+N]})$. The design idea behind this is that when predicting the next code's position, the model should not only be aware of previous steps' codes and their position information but also should be aware of \emph{current} code information. The negative log-likelihood (NLL) for position autoregressive training is: 

\begin{equation}
    \mathcal{L}_{\text{position}} = \mathbb{E}[-\log p(\overline{\mathbf{P}}_l | \overline{\mathbf{M}}_{\leq l}, \overline{\mathbf{P}}_{<l})].
\end{equation}

\textbf{Training $\&$ Inference.} The total loss for training Stackformer is defined as:

\begin{equation}
    \mathcal{L} = \mathcal{L}_{\text{code}} + \mathcal{L}_{\text{position}}.
\end{equation}
The inference procedure is illustrated in Algorithm \ref{alg:sample}, where we take the unconditional generation as an example and conditional generation can be derived accordingly. 

\begin{algorithm}[!h]
    \caption{Unconditional sampling of Stackformer.}
    \label{alg:sample}
    \renewcommand{\algorithmicrequire}{\textbf{Input:}}
    \renewcommand{\algorithmicensure}{\textbf{Output:}}
    \begin{algorithmic}[1]
        \REQUIRE 
        The sample step $N$; \\
        The code sequence $M_{sample}$ of a single $<$sos$>$; \\
        The code position sequence $P_{sample}$ of a single $<$sos$>$. \\
        \ENSURE The generated image $\mathcal{I}$.    
        \FOR{each $n \in [1,N]$}
            \STATE $H_c = \operatorname{Code-Transformer}(M_{sample}, P_{sample})$;
            \STATE Sample next code $M_n$ by $H_c$;
            \STATE $M_{sample} = \operatorname{concat}(M_{sample}, M_n)$;
            \STATE $H_p = \operatorname{Position-Transformer}(H_c, M_{sample}[2:])$;
            \STATE Mask already sampled positions to avoid conflicts;
            \STATE Sample next code position $P_n$ by $H_p$;
            \STATE $P_{sample} = \operatorname{concat}(P_{sample}, P_n)$;
        \ENDFOR
        \STATE Re-map $M_{sample}$ to 2D code map according to $P_{sample}$ and the rest are filled with the mask code;
        \STATE Decode the code map to the image $\mathcal{I}$
        \RETURN The generated image $\mathcal{I}$
    \end{algorithmic}
\end{algorithm}

\begin{figure*}
  \centering
  \includegraphics[width=1.0\linewidth]{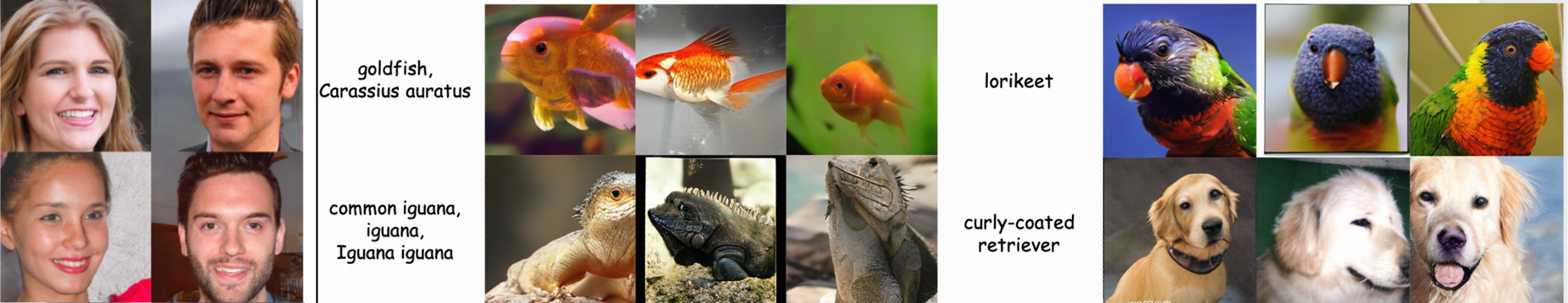}
  \caption{Left: Our unconditional generated images on FFHQ benchmark. Right: Our class-conditional generated images on ImageNet.}
  \label{imagenet_visual}
\end{figure*}

\section{Experiment}

\subsection{Experimental Settings}

\textbf{Benchmarks.} We validate our model for unconditional, class-conditional, and text-conditional image generation tasks on FFHQ\cite{karras2019style}, ImageNet\cite{deng2009imagenet}, and MS-COCO\cite{lin2014microsoft} benchmarks respectively, with $256 \times 256$ image resolution.

\textbf{Metrics.} Following previous works\cite{esser2021taming, lee2022autoregressive, yu2021vector}, the standard Frechet Inception Distance (FID)\cite{heusel2017gans} is adopted for evaluating the generation and reconstruction quality (denoted as rFID). Inception Score (IS)\cite{barratt2018note} is also used for class-conditional generation on the ImageNet benchmark. FID and IS are calculated by sampling 50k images. rFID is calculated over the entire test set.

\textbf{Implemented details.} The architecture of MQ-VAE exactly follows \cite{esser2021taming} except for the proposed mask and de-mask modules, with the codebook size of $K=1024$. For the de-mask module, the sub-module number $H=8$. For the Stackformer, we implement two settings: a small version uses $18$ transformer encoder blocks for the Code-Transformer and another $6$ transformer encoder blocks for the Position-Transformer with a total of $307$M parameters, and a base version uses $36$ transformer encoder blocks for the Code-Transformer and another $12$ transformer encoder blocks for the Position-Transformer with a total of $607$M parameters to further demonstrate our scalability. The generation results are reported with a 25\% mask ratio at $32 \times 32$ resolution feature map using eight RTX-3090 GPUs. Top-k and top-p sampling are used to report the best performance. More details could be found in the supplementary.

\subsection{Comparison with state-of-the-art methods}

\begin{table}[h]
    \centering
    \small
    \begin{tabular}{l c c}
    \toprule[1pt]
    Methods & \#Params & FID$\downarrow$ \\
    \hline
    DCT\cite{nash2021generating} & 738M & 13.06 \\
    VQGAN\cite{esser2021taming} & (72.1+307)M & 11.4 \\
    RQ-Transformer\cite{lee2022autoregressive} & (100+355)M & 10.38 \\
    Mo-VQGAN\cite{zheng2022movq} & (82.7+307)M & 8.52 \\
    \hline
    \textbf{Stackformer} & (44.4+307)M & \textbf{6.84} \\
    \textbf{Stackformer} & (44.4+607)M & \textbf{5.67} \\
    \toprule[1pt]
    \end{tabular}
    \caption{Comparison of autoregressive unconditional generation at million-level parameters on FFHQ\cite{karras2019style} benchmark. \#Params splits in (VAE + autoregressive model).}
    \label{tab:unconditional_million_autoregressive}
\end{table}

\begin{table}[]
    \centering
    \small
\begin{tabular}{l c c c c}
    \toprule
    Model Type & Methods & \#Params & FID$\downarrow$ & IS$\uparrow$ \\
    \hline
    Diffusion & ADM\cite{dhariwal2021diffusion} & 554M & 10.94 & 101.0 \\
    \hline
    GAN & BigGAN\cite{brock2018large} & 164M & 7.53 & 168.6 \\
    GAN & BigGAN-deep\cite{brock2018large} & 112M & 6.84 & 203.6 \\
    \hline
    Bidirection & MaskGIT\cite{chang2022maskgit} & 227M & 6.18 & 182.1 \\
    \hline
    \scriptsize Autoregressive & DCT\cite{nash2021generating} & 738M & 36.5 & n/a \\
    \scriptsize Autoregressive & VQ-GAN$\dagger$\cite{esser2021taming} & 679M & 17.03 & 76.85 \\
    \scriptsize Autoregressive & \footnotesize RQ-Transformer\cite{lee2022autoregressive} & 821M & 13.11 & 104.3 \\
    \scriptsize Autoregressive & Mo-VQGAN\cite{zheng2022movq} & 383M & 7.13 & 138.3 \\
    \hline
    \scriptsize Autoregressive & \textbf{Stackformer} & 651M & \textbf{6.04} & \textbf{172.6} \\
    \bottomrule
\end{tabular}
\caption{Comparison of class-conditional image generation at
million-level parameters without rejection sampling on ImageNet\cite{deng2009imagenet}. $\dagger$ denotes the model we train with the same setup with ours.}
\label{tab:class_conditional_million}
\end{table}

\textbf{Unconditional generation.} We first compare with million-level state-of-the-art autoregressive models in Table \ref{tab:unconditional_million_autoregressive}. Our model significantly outperforms other autoregressive models with the same parameters (307M). With more parameters, we further increase the FID from 6.84 to 5.67, which demonstrates our scalability. We also compare with other types of unconditional state-of-the-art and large-scale big models in Table \ref{tab:unconditional_other}, where we also achieve top-level performance. Our qualitative unconditional generation results are shown on the left of Figure \ref{imagenet_visual}.


\begin{table}[h]
    \centering
    \small
    \begin{tabular}{l c c c}
    \toprule[1pt]
    Model Type & Methods & \#Params & FID$\downarrow$ \\
    \hline
    VAE & VDVAE\cite{child2020very} & 115M & 28.5 \\
    Diffusion & ImageBART\cite{esser2021imagebart} & 3.5B & 9.57 \\
    GAN & StyleGAN2\cite{karras2020analyzing} & - & 3.8 \\
    GAN & BigGAN\cite{brock2018large} & 164M & 12.4 \\
    Autoregressive & ViT-VQGAN\cite{yu2021vector} & 2.2B & 5.3 \\
    \hline
    Autoregressive & \textbf{Stackformer} & 651M & \textbf{5.67} \\
    \toprule[1pt]
    \end{tabular}
    \caption{Comparison with other types of state-of-the-art generative models and large-scale \emph{billion-level} parameters autoregressive models on unconditional FFHQ\cite{karras2019style} benchmark.}
    \label{tab:unconditional_other}
\end{table}

\begin{table}[]
    \centering
    \small
\begin{tabular}{l c c c c}
    \toprule
    Model Type & Methods & \#Params & FID$\downarrow$ & IS$\uparrow$ \\
    \hline
    Diffusion & ImageBART\cite{esser2021imagebart} & 3.5B & 21.19 & 61.6 \\
    \scriptsize Autoregressive & VQVAE2\cite{razavi2019generating} & 13.5B & ~31 & ~45 \\
    \scriptsize Autoregressive & VQ-GAN\cite{esser2021taming} & 1.4B & 15.78 & 78.3 \\
    \scriptsize Autoregressive & ViT-VQGAN\cite{yu2021vector} & 2.2B & 4.17 & 175.1 \\
    \scriptsize Autoregressive & \footnotesize RQ-Transformer\cite{lee2022autoregressive} & 3.8B & 7.55 & 134 \\
    \hline
    \scriptsize Autoregressive & \textbf{Stackformer} & 651M & \textbf{6.04} & \textbf{172.6} \\
    \bottomrule
\end{tabular}
\caption{Comparison between our model and large-scale \emph{billion-level} parameters models of class-conditional generation without rejection sampling on ImageNet\cite{deng2009imagenet} benchmark.}
\label{tab:class_conditional_billion}
\end{table}

\begin{figure*}
  \centering
  \includegraphics[width=1.0\linewidth]{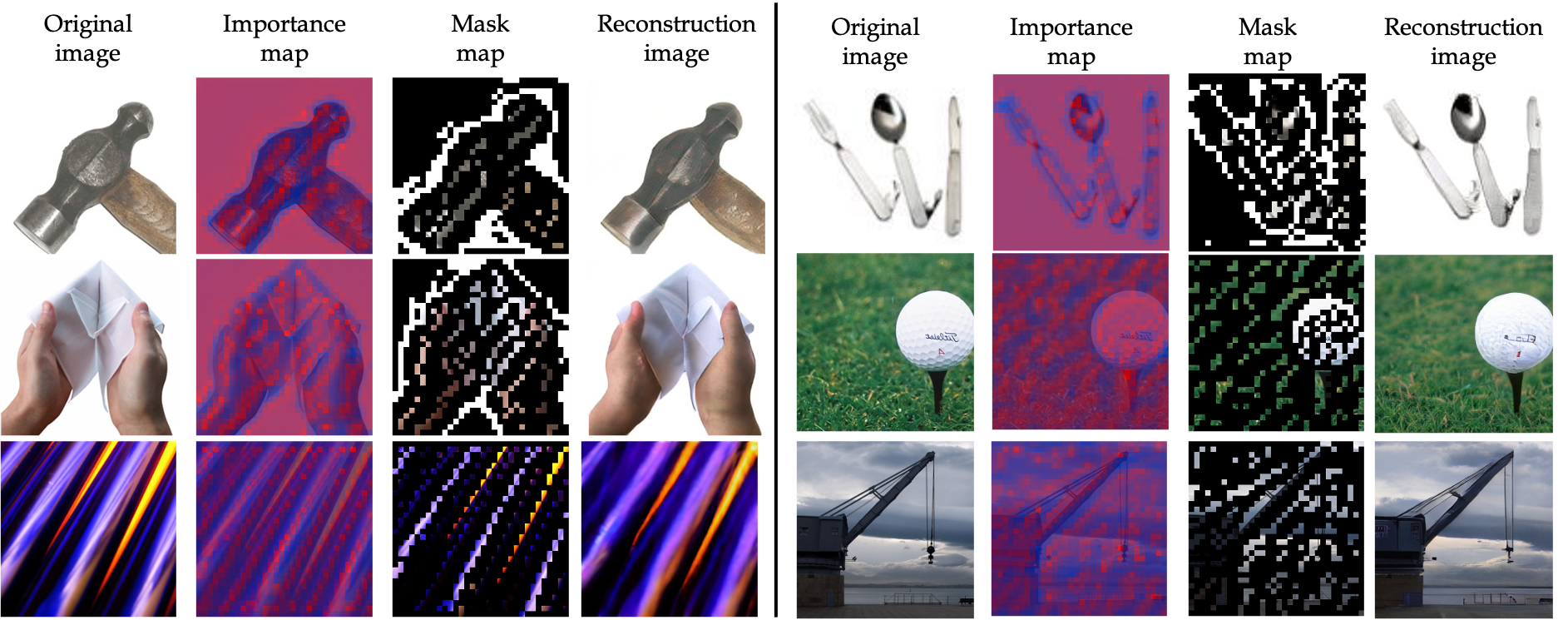}
  \caption{The visualization of our adaptive mask module which learns to mask unimportant regions on ImageNet\cite{deng2009imagenet}. In the importance map, red denotes high scores while blue denotes low scores.}
  \label{mask}
\end{figure*}

\textbf{Class-conditional generation.} We first compare with million-level state-of-the-art in Table \ref{tab:class_conditional_million}. We achieve the best FID score compared to all types of models including the recent Mo-VQGAN\cite{zheng2022movq} and RQ-Transformer\cite{lee2022autoregressive}. We also compare our \emph{million-level} model with existing \emph{billion-level} big models in Table \ref{tab:class_conditional_billion}, where we also achieve top performance with fewer parameters and is only inferior to ViT-VQGAN big model. Our qualitative class-conditional generation results are shown on the right of Figure \ref{imagenet_visual}.


\begin{table}[]
    \centering
    \footnotesize
    \begin{tabular}{l c c}
        \toprule
        Model Type & Method & FID$\downarrow$  \\
        \hline
        GAN & DMGAN\cite{zhu2019dm} & 32.64 \\
        GAN & XMCGAN$\dagger$\cite{zhang2021cross} & 50.08 \\
        GAN & DFGAN\cite{tao2020df} & 21.42 \\
        GAN & SSA-GAN\cite{liao2022text} & 19.37 \\
        GAN & DSE-GAN\cite{huang2022dse} & 15.30 \\
        Diffusion & VQ-Diffusion \cite{gu2021vector} & 19.75 \\
        Autoregressive & VQ-GAN$\dagger$\cite{esser2021taming} & 22.28 \\
        \hline
        Autoregressive & \textbf{Stackformer} & \textbf{10.08} \\
        \bottomrule
        \end{tabular}
    \caption{Comparison of text-conditional generation on MS-COCO\cite{lin2014microsoft} without using extra web-scale data or pre-trained models. \\
$\dagger$ denotes reproduced results under our same experimental setting.}
\label{tab:text_conditional}
\end{table}

\textbf{Text-conditional generation.} We compare with existing text-conditional state-of-the-art without extra web-scale data or pretrained models on MS-COCO\cite{lin2014microsoft} for fair comparison in Table.\ref{tab:text_conditional}. We achieve 18.6\% FID improvement.

\begin{figure}[H]
  \centering
  \includegraphics[width=1.0\linewidth]{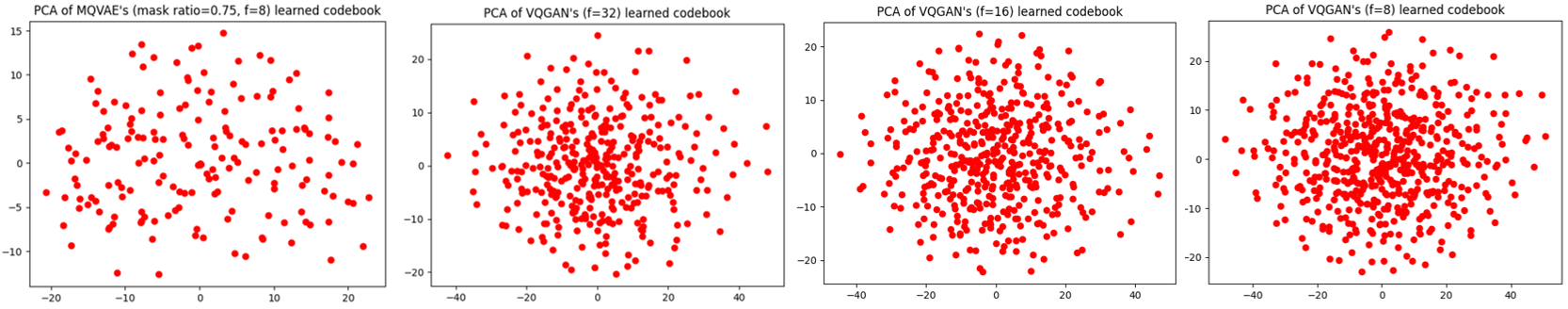}
  \caption{The PCA of learned codebook (1024 codebook size).}
  \label{fig_pca}
\end{figure}


\subsection{Ablations}

We conduct ablations on $16\times16$ resolution feature map using four RTX-3090 GPUs for saving computation resources and the experimental trends are all consistent with $32\times32$ resolution feature map of the main results.

\begin{table}[]
    \centering
    \small
    \begin{tabular}{c c c c c c}
    \toprule[1pt]
    mask ratio & $f$ & mask type & rFID$\downarrow$ & FID$\downarrow$ & usage$\uparrow$ (\%) \\
    \hline
    0\% & 32 & - & 8.1 & 13.5 & 70.02 \\
    0\% & 16 & - & 4.46 & 11.4 & 63.89 \\
    \hline
    10\% & 16 & adaptive & 4.55 & 7.81 & 72.34 \\
    25\% & 16 & adaptive & 4.79 & 7.67 & 78.22 \\
    25\% & 16 & random & 6.13 & 12.21 & 67.48 \\
    50\% & 16 & adaptive & 5.31 & 8.36 & 84.29 \\
    50\% & 16 & random & 7.855 & 15.67 & 69.04 \\
    75\% & 16 & adaptive & 7.62 & 11.71 & 87.60 \\
    75\% & 16 & random & 10.58 & 17.62 & 70.41 \\
    \toprule[1pt]
    \end{tabular}
    \caption{Ablations of adaptive mask module on FFHQ. Here $f$ is the downsampling factor. The codebook usage is calculated as the percentage of used codes over the entire test dataset.}
    \label{tab:ablation_adaptive_mask_module}
\end{table}

\textbf{Ablations of adaptive mask module.} As shown in Table \ref{tab:ablation_adaptive_mask_module}, our proposed learned adaptive mask mechanism significantly outperforms the random one, which quantitatively validates that the adaptive mask module enables the model to learn perceptually important regions. 

Our model with 10\% and 25\% mask radio has only slightly inferior reconstruction compared with VQGAN\cite{esser2021taming}, but achieves a significant improvement in generation quality, which indicates that the effectiveness of focusing autoregressive models on modeling important regions. When we further increase the mask radio to 50\% and 75\%, the final generation quality drops, we believe the reason lies that an improper high mask radio will inevitably mask some important regions that greatly decrease the reconstruction results and hinder autoregressive modeling.

The redundancy of the existing learned codebook can be verified from two aspects: i) the PCA of the learned codebook in Figure \ref{fig_pca}, where each point is a code and closer codes have more similar semantics. We show many codes in VQGAN's codebook overlap, which indicates these codes have nearly the same semantics and are thus redundant. The redundancy increase (more overlaps) when VQGAN uses more code to represent images (smaller downsampling factor $f$). The redundancy is largely alleviated in our MQ-VAE. ii) in Table \ref{tab:ablation_adaptive_mask_module}, a higher codebook usage indicates more ``useful" codes in the codebook and thus less redundant. VQGAN has a lower usage compared with our MQ-VAE.

\begin{figure}
  \centering
  \includegraphics[width=1.0\linewidth]{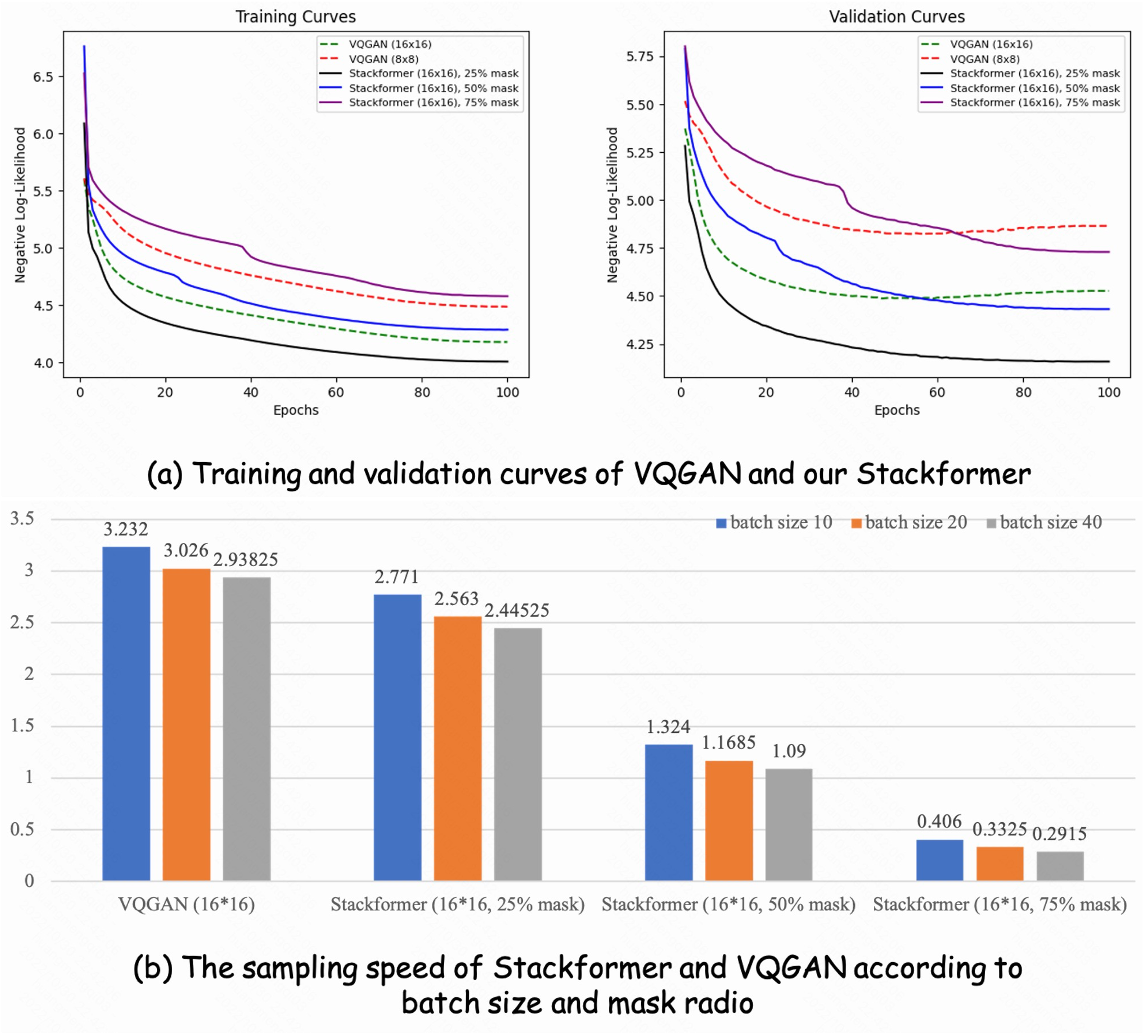}
  \caption{Comparison of training $\&$ validation curves and sample speed between VQGAN\cite{esser2021taming} and Stackformer.}
  \label{curves_speed}
\end{figure}

We visualize the training and validation curves of VQGAN and Stackformer in Figure \ref{curves_speed}(a). Previous autoregressive models\cite{esser2021taming, lee2022autoregressive, yu2021vector} always suffer from the overfitting problem while our Stackformer successfully gets rid of it, which indicates the better generalization of our masked discrete representation and our model.

We compare the sampling speed on a single RTX-1080Ti in Figure \ref{curves_speed}(b). Compared
with VQGAN, our 25\% mask radio model achieves 32.72\% quality improvement and 15.45\% speed improvement, while Our 50\% mask radio model achieves 26.67\% quality improvement and 61.1\% speed improvement. Therefore, our design enables a flexible trade-off between speed and quality.

Finally, We visualize the learned mask in Figure \ref{mask}, with a 75\% mask ratio on $32\times32$ resolution feature map, which validates that our proposed adaptive mask mechanism successfully learns to preserve the perceptual important image regions, \emph{i.e.}, the structural and edge regions of objects.

\begin{table}
	\centering
	\footnotesize
	\begin{tabular}{c c c}
    \toprule
    Model setting & rFID$\downarrow$ & FID$\downarrow$ \\
    \hline
    VQGAN & 4.46 & 11.4 \\ 
    \hline
    VQGAN* & 4.17 & 11.02 \\
    \hline
    MQ-VAE w/o de-mask & 7.02 & 10.74 \\
    \hline
    MQ-VAE w/ de-mask (SA) & 6.56 & 9.8 \\
    \hline
    MQ-VAE w/ de-mask (DC-SA w/o mask update) & 5.84 & 8.92 \\
    \hline
    MQ-VAE w/ de-mask (DC-SA w/ mask update) & 5.31 & 8.36 \\
    \bottomrule
\end{tabular}
\caption{Ablations of adaptive de-mask module on FFHQ. SA for self-attention and DC-SA for direction-constrained self-attention. ``VQGAN*'' is the stronger baseline, where the same numbers of SA and Resnet blocks as MQ-VAE's de-mask module are added.}
\label{ablations_demask}
\end{table}

\textbf{Ablations of adaptive de-mask module.} In Table \ref{ablations_demask}, we show that MQ-VAE outperforms VQGAN and the stronger baseline (``VQGAN*''), which validates our effectiveness. We could conclude that the proposed direction-constrained self-attention and the mask updating mechanism both improve the reconstruction and generation quality.




\section{Conclusion}

In this study, we point out that the existing two-stage autoregressive generation paradigm ignores distinguishing the perceptual importance of different image regions, which brings redundancy that not only degrades generation quality but also decreases generation speed. We propose a novel two-stage generation paradigm with MQ-VAE and Stackformer to relieve the model from redundancy. MQ-VAE incorporates the \emph{adaptive mask module} to mask redundant region features before quantization and the \emph{adaptive de-mask module} to recover the original feature map after quantization. Stackformer then efficiently predict the combination of both codes and their positions. Comprehensive experiments on various types of image generation tasks validate the effectiveness and efficiency of our method. 


\section{Acknowledgments}
This work is supported by National Natural Science
Foundation of China under Grants 62222212 and U19A2057, Science Fund for Creative Research Groups under Grant 62121002.

{\small
\bibliographystyle{ieee_fullname}
\bibliography{egbib}
}

\end{document}